\documentclass[conference,compsoc]{IEEEtran}
\usepackage{amsmath}

\ifCLASSOPTIONcompsoc
  \usepackage[nocompress]{cite}
\else
  \usepackage{cite}
\fi
\ifCLASSINFOpdf
\else
\fi
\begin{document}
%
\title{The Challenge of Imputation in Explainable Artificial Intelligence Models}

\author{\IEEEauthorblockN{Muhammad Aurangzeb Ahmad\IEEEauthorrefmark{1},
Carly Eckert\IEEEauthorrefmark{1}, Ankur Teredesai\IEEEauthorrefmark{1}\IEEEauthorrefmark{2} }\\
\IEEEauthorblockA{
\IEEEauthorrefmark{1}KenSci Inc, Seattle, WA\\
\IEEEauthorrefmark{2}Department of Computer Science, University of Washington, Tacoma\\
\{muhammad, carly, ankur\}@kensci.com
}}


%


\maketitle


\begin{abstract}
Explainable models in Artificial Intelligence are often employed to ensure transparency and accountability of AI systems. The fidelity of the explanations are dependent upon the algorithms used as well as on the fidelity of the data. Many real world datasets have missing values that can greatly influence explanation fidelity. The standard way to deal with such scenarios is imputation. This can, however, lead to situations where the imputed values may correspond to a setting which refer to counterfactuals. Acting on explanations from AI models with imputed values may lead to unsafe outcomes. In this paper, we explore different settings where AI models with imputation can be problematic and describe ways to address such scenarios.
\end{abstract}

\section{Introduction}
Even though the field of Artificial Intelligence is more than sixty years old, it is only in the last decade or so that AI systems are being increasingly interwoven into the fabric of the socio-technical apparatus of the society and are thus having a massive impact on society. This increasing incorporation of AI has led to increased calls for accountability and regulation of AI systems \cite{gunning2017explainable}. Model explanations are considered to be one of the most important ways to provide accountability of AI systems. The model explanations, however, can only be as good as the data on which the algorithms are based. This is where the issue of missing and imputed data becomes pivotal for model explanations. In some domains like healthcare, almost all datasets have missing values \cite{cismondi2013missing}. As many applications of AI in healthcare are patient-oriented, decisions that are informed by AI and ML models can potentially have significant clinical consequences. Additionally, the requirements for a responsible and robust AI solution in critical domains like healthcare, criminal justice system etc. require additional considerations that affect safety and compliance with regulations \cite{ahmad2018interpretable}.

One of the main reasons why imputation is used in AI and machine learning models is that it allows the use of all available data for model building instead of restricting oneself to a subset of completed records, likely achieve via deletion. The net effect is that models built using imputed data often result in better predictive performance. Consequently, imputation may be necessary if performance is more important than explainability. Even outside of data missingness, researchers acknowledge that there is a trade-off between explainable AI models and their predictive performance \cite{weller2017challenges} \emph{i.e.,} in general the more explainable a model is the less predictive power it will have and vice versa. Thus, Deep Learning models, which are epitome of black box models, are relatively opaque but models like linear regression models are more straightforward to interpret. The choice of the underlying algorithm for model explanation is often dependent upon the risk involved \cite{ahmad2018interpretable}.  For example, in a problem predicting the likely need for palliative care, the physician must know why the prediction is being made since the prediction may changes the plan of care for a patient. In this scenario predictive performance could be sacrificed for intelligibility of the prediction. On the other hand, if the application is to predict the upcoming patient census in an emergency department the ED charge nurse may be less concerned with model explainability. 

Concerns regarding the safety of AI systems become paramount when we consider that incorrect imputation can lead to explanations that may not have fidelity with the underlying phenomenon and thus may lead to potentially harmful actions, especially in decision support systems. Consequently, questions regarding imputation are of paramount importance from a safety and even regulatory perspective in high risk fields like healthcare, criminal justice system, and elsewhere. In this paper we will explore multiple scenarios that can arise when dealing with imputation in model explanations in AI. To illustrate the problems with imputation, we consider use cases in the healthcare domain using data from a hospital system in the mid-Western United States consisting of data from 76 thousand patients, similar but not identical to the dataset used in \cite{ahmad2018death}.

\section{Imputation with Unsafe Outcomes}
It is important to note that explainability is just one mechanism to ensure safety of AI systems and to mitigate against unwarranted and unforeseen outcomes. Optimizing for explinability via post-hoc models while simultaneously optimizing for performance via imputation may lead to unsafe outcomes as we demonstrate with examples from healthcare and the criminal justice system.

\subsection{Use Case: Imputation and Patient Safety}
One of the main dangers for using imputed values for model explanations is that the explanations that are produced may in fact have factors which are absurd or do not make domain sense. To illustrate, consider the following example based off of the problem of predicting the length of stay of a patient at a hospital at the time of admission. Suppose that a black box model is used (e.g., Extreme Gradient Boosting) as the underlying predictive model. One way to extract explanations from this models is to use post-hoc models like LIME \cite{ribeiro2016should} or Shapley Values \cite{shapley1953value}. To illustrate such a case, consider the output for model explanation in Table \ref{tab:lime} for predicting length of stay in a hospital using the LIME model. The Table gives the factor and its associated value, as well as the relative importance of the factor for the explanation and whether it was imputed or not. The third most important factor, the albumin level which refers to a lab test, is an imputed value. 

Suppose the end user for this system is a clinician who is trying to understand why the predictions are being made. To get the full context she looks at her patient's lab results (\emph{e.g.,} a list of the most recent lab tests and results for the patient.) She will observe a missing value for Albumin but when she looks at the explanation for the predictions and sees the lab value for a lab which was not actually performed. The problem with this scenario is that it could cast doubt on the entire machine learning solution in the mind of the user since the explanation for the prediction is at odds with the patient history. Consider the alternative ways in which this scenario can play out \emph{e.g.,} if the explanation for the prediction is that the patient has a low readmission risk due to a low Troponin. Troponin refers to a group of proteins found in skeletal and heart muscle fibers that are responsible for regulating muscular contraction \cite{ebashi1971troponin}. However if Troponin has not been evaluated, the physician may overlook that fact and assume that an acute coronary event has already been ruled out. However if the acute coronary event is indeed the real reason for the patient is at risk then it is may lead to a catastrophic outcome. 

Additionally imputing from a biased population (hospitalized patients, for example) does not give one a true representation of the value in a patient without pathology \emph{e.g.,} Troponin is only measured in people with an expected heart attack or other acute cardiac issue. Therefore, the distribution of values for Troponin does not represent that of the non-diseased population. If a 20 year old is in the Emergency Department for abdominal pain from appendicitis, one should not impute a Troponin. What this example illustrates is that in some contexts there is additional context which may not be captured by data and imputing without domain knowledge can be potentially hazardous.

\begin{table}
\begin{tabular}{llll}
\hline
\textbf{Feature}  & \textbf{Value} & \textbf{Importance} & \textbf{Imputed} \\
\hline
Readmits last year       & 2  & 0.67 & 0     \\
Creatinine Level        & 1.6  & 0.54 & 0      \\
Albumin Level   & 2.4  & 0.36 & 1     \\
Avg. Cigarettes/week   & 28  & 0.18 & 0     \\
Systolic BP   & 90  & 0.17 & 1   \\
\hline
\end{tabular}
\caption{LOS Related Explanation Factor for a patient}
\label{tab:lime}
\end{table}

\subsection{Use Case: Imputation in Criminal Justice System}
Another use case where imputation can lead to adverse consequences is in the criminal justice system. If the prediction task is to predict if the person is likely to re-offend or not, then imputing values may lead to scenarios where the person is deemed low risk because variables related to the person's risk profile are imputed based on people who are similar to him in the larger population. The problem with this approach using imputation is that if the sample is biased against minority populations then imputation may always leads to imputing negative tendencies and thus higher risk scores for the minority population. This may happen even if demographic features are not used in model building \cite{berk2018fairness} and because such factors are not used they will also not show up as explanation factors. The end user will only see that the person is a high risk because of certain, possibly violent, tendencies or past history. If the imputed nature of such explanations is not highlighted then this is likely to lead to faulty recommendations and even wrongfully long sentences.

\section{Alternates to Imputation in Model Explanations}
Since model performance is a requirement in most machine learning applications and explanation fidelity is needed, it is important to consider alternatives to imputation. One way to address the problem of missing values in some domains is to use indicator variables because the presence and absence of certain variable is still useful information. In Electronic Health Records (EHR) datasets majority of the labs and vitals are missing for most patients because testing for all labs is unnecessary and thus the labs were never ordered. Most machine learning predictive models infer the missing values via imputation. If a result exists for a lab then that implies that the ordering clinician deemed it necessary. Therefore, the fact that the lab was ordered is itself useful information. We propose that this information can be incorporated in machine learning models in at least two ways. In the first scheme, indicator variables for features like labs and vitals can be used.  Instead of the original values for the variables we record their presence and absence as follows:
\begin{equation}
  \begin{aligned}
    f(x)=
    \begin{cases}
      0, &\textit{if}\ x=null \\
      1, &\textit{otherwise}
    \end{cases}
  \end{aligned}
\end{equation}
The alternate method is to use central values \textit{i.e.,} average, median, truncated average etc. or central values by cohort for variables like labs and vitals as follows:
\begin{equation}
    f(x)=
    \begin{cases}
      norm(x), ~\textit{if}\ x=null \\
      x, ~\textit{otherwise}
    \end{cases}
\end{equation}
  
It should be noted that even though one can use imputed variables instead of the corresponding variable with the actual and missing values, in most cases the expected performance of a model based on the indicator variables would be less as compared to a model with imputed values. Consider Decision Trees where the split of the decision tree branch needs to be decided for a variable in integer space \emph{e.g.,} age. A popular scheme to do this is to consider the midpoints between actual data points and determine which splits give the best information gain or other information theoretic criteria.

Assuming that most variable spaces will be non-binary spaces, the split that one gets from the indicator variables will be different since the loss of entropy given by the equation would be different. In other words the models which are created from the imputed variables vs. the indicator variables would be different. To illustrate this, we build multiple prediction models on the mid-Western EHR data described above. The results of prediction for the models are given in Table \ref{tab:results}. The model with no missing values refers to a model that was built with only using variables with no missing values. We note that even for the model which used variables with missing values, we did not use a variable where more than fifty percent of the values were missing. Since the space of such variables is much smaller as compared to the space considered for all the variables, the performance of the models is quite mediocre \emph{i.e.,} an MAE of 4.09 days. The results for the model with a sophisticated imputation technique like MICE are quite good with an MAE of 2.08 days and that for the model with indicator variables is 2.46 days. 

Lastly, we also tried model building with average values and the results are better than what we get for indicator variables model but with the caveat that these values do not correspond to anything that was measured but rather the values are inferred for normalcy. While the indicator variables models is not as good as the imputation model, it may still be sufficiently good to be used in a production setting. The added advantage of using such a model would be that the model explanations would actually correspond to how the data is \emph{i.e.,} indicate the presence or absence of a variable.

\begin{table}
\begin{tabular}{lll}
\hline
\textbf{Model}  & \textbf{MAE} & \textbf{MSE} \\
\hline
Model with no Missing Variables      & 4.09 &  213.07 \\
Model with MICE Imputation        & 2.08 & 13.29 \\
Model with Indicator Variables  & 2.46 & 18.05   \\
Model with Averages Imputed  & 2.40 & 17.80   \\
\hline
\end{tabular}
\caption{Results Summary for LOS Models in days}
\label{tab:results}
\end{table}

\section{Explanability vs. Types of Data Missingness}
All imputation methods add bias to the data which in turn affects the fidelity of the model built based on the imputed data.  This implies that model fidelity is dependent upon the nature of the missing data and the imputation technique. Data can be, Missing at random, not missing at random and completely missing at random \cite{little2019statistical}. Consider how each of these missingness types affect model explanations: If the data is missing completing at random (MCAR) then even though imputation adds bias to the data, the introduced bias would be local \emph{i.e.,} at the instance level and not global since the data is not systematically missing. On the other hand, if the data is missing at random (MAR) then it means that the missing data is at least partially dependent upon other variables in the data \emph{e.g.,} in a survey people who are in the service sector may be less likely to report their income, clinicians may not order certain labs to be tested if they think that the patient is less likely to have certain conditions. Imputing in such cases, add bias to the data at the global level.

Imputation for data that is missing at random can be done by employing domain knowledge or some other known aspect of data \emph{e.g.,} expected values for populations with normal characteristics may be imputed for some laboratory tests if the patients do not have certain conditions. Lastly, if the data is not missing at random (NAMR) then sophisticated statistical methods like MICE, 3D-MICE are often used used for imputation. MICE or multiple imputation by chained equations \cite{azur2011multiple} is a method where each missing variable is imputed by a separate model but it maintains consistency between imputations by means of passive imputation. It should be noted that however, all imputation methods introduce biases and studies have shown that imputed values can be wildly off from real values \cite{luo20173d}. If the values are off then the explanations will be incorrect. Thus, explanations for AI models are less likely to be off in case of the data missingness is because of MAR or MCAR but imputation for NMAR data could occasionally lead to explanations that may not correspond to how the model actually works. We address these and other concerns in the section on operationalization of explanations. 

Implementations of tree based models like XGBoost handle missing values by having a default direction for nodes with missing values in the current instance set \cite{chen2016xgboost}. The direction of the missing data is thus used as a proxy of data missingness. Such implementations of tree based algorithms minimize error loss in the training phase. However this also means that if the distribution of missing values in the test set is different from that of the training set then the performance of the predictive models will suffer and consequently the quality of explanations generated by the models would also degrade.

\section{Operationalizing Explanations with Imputation}
An important axis of accountability of Artificial Intelligence systems is Fairness. There are multiple notions of what constitutes fairness in machine learning \cite{srivastava2019mathematical}, to illustrate the effect of imputation on fairness and explainability we focus on group level notions of fairness. A system that is fair need not be explainable if the underlying algorithm can ensure that the various groups of interest are being scored in a fair manner even when the model is blackbox. Requiring the model to be explainable and interpretable may require it impute values about groups which could render the model unfair since imputed values can add bias against certain protected groups \emph{e.g.,} ethnicity, race, gender etc. The choice of the trade-off should be made with respect to the use case and the application domain.

It has been noted that generating and operationalizing explanations is not strictly a machine learning problem \cite{weller2017challenges}. The same machine learning model may be used to generate disparate explanations if the use case or the end users are different \cite{ahmad2018interpretable}. The risk profile of the end user determines how explanations should be operationalized. If the use of explainable models is an absolute requirement in a setting where there are a large number of missing values then appropriate disclaims regarding data missing and imputation should be given to the end user. Even with such messaging it is still possible that automation bias can creep in and lead to unsafe recommendations and outcomes \cite{mullainathan2017does}. Additionally, imputation of missing values can lead to models which are unfair and thus require special attention \cite{martinez2019fairness}.

\section{Conclusion}
In this paper we have highlighted a number of issues related to extracting explanations from AI and machine learning models that use imputation of missing data. If such issues are not surfaced to the end user then acting upon explanations from such models can lead to dire consequence in terms of human safety and well being. The case of incorrect explanations in the healthcare domain is especially acute. At the minimum use of explanations at the local instance level should be accompanied by appropriate disclaimers and end users should be properly educated about the potential hazards of incorrect explanations. The other alternative is to take an engineering approach to explanations and assume that assurance of AI systems takes precedence over explainability. Just as one's choice of algorithm for prediction problems needs to take into account the risk profile of the prediction, the same should also apply for imputation.

\end{document}